\documentclass{article}
\usepackage{ijcai16}
\usepackage{graphicx}
\usepackage[pdftex]{hyperref}
\usepackage{multirow}
\usepackage{listings}

\usepackage{times}
\usepackage{amsmath}
\usepackage{amssymb}

\title{Deep, Convolutional, and Recurrent Models for Human Activity Recognition using Wearables}
\author{Nils Y.\ Hammerla$^{\mathbf{1,2}}$, Shane Halloran$\mathbf{^2}$, Thomas Pl{\"o}tz$\mathbf{^2}$ \\
$^1$babylon health, London, UK \\
$^2$Open Lab, School of Computing Science, Newcastle University, UK  \\
nils.hammerla@babylonhealth.com}

\begin{document}

\maketitle

\begin{abstract}
Human activity recognition (HAR) in ubiquitous computing is beginning to adopt deep learning to substitute for well-established analysis techniques that rely on hand-crafted feature extraction and classification techniques. From these isolated applications of custom deep architectures it is, however, difficult to gain an overview of their suitability for problems ranging from the recognition of manipulative gestures to the segmentation and identification of physical activities like running or ascending stairs. In this paper we rigorously explore deep, convolutional, and recurrent approaches across three representative datasets that contain movement data captured with wearable sensors. We describe how to train recurrent approaches in this setting, introduce a novel regularisation approach, and illustrate how they outperform the state-of-the-art on a large benchmark dataset. Across thousands of recognition experiments with randomly sampled model configurations we investigate the suitability of each model for different tasks in HAR, explore the impact of hyperparameters using the fANOVA framework, and provide guidelines for the practitioner who wants to apply deep learning in their problem setting.
%Human activity recognition (HAR) in ubiquitous computing is beginning to adopt deep learning to substitute for well-established analysis techniques that rely on hand-crafted feature extraction and classification techniques for time-series data collected with wearable movement sensors like accelerometers or gyroscopes. From isolated applications of custom deep architectures it is, however, difficult to obtain an overview of their suitability for problems ranging from the recognition of manipulative gestures to the segmentation and identification of physical activities like running or ascending stairs. In this paper we rigorously explore deep, convolutional, and recurrent approaches across three representative data-sets. We report results from thousands of experiments with randomly sampled hyper-parameters and provide an unbiased comparison of the strengths and weaknesses of the different models. We describe in detail how to train these models and investigate the effect of individual hyperparameters using the f-anova framework. We illustrate how recurrent models enable novel applications in HAR and show that their performance exceeds that of the best performing systems in the domain. We conclude in guidelines for parameter exploration and model selection that will help practitioners maximise the benefit that deep learning offers in this domain.
\end{abstract}

\section{Introduction}

Deep learning represents the biggest trend in machine learning over the past decade. Since the inception of the umbrella term the diversity of methods it encompasses has increased rapidly, and will continue to do so driven by the resources of both academic and commercial interests. Deep learning has become accessible to everyone via machine learning frameworks like Torch7 \cite{Collobert_NIPSWORKSHOP_2011}, and has had significant impact on a variety of application domains \cite{lecun2015deep}.

%One field that has yet to benefit of deep learning is Ubiquitous Computing (ubicomp). The majority of recognition systems in ubicomp aimed at the analysis of human movement data rely on traditional approaches to time-series analysis, where multivariate time-series data is first segmented, then subjected to a manually designed feature extraction process [], which forms the basis for well-established learning techniques like decision trees [] or instance-based methods like k-nearest neighbour [].

One field that has yet to benefit of deep learning is Human Activity Recognition (HAR) in Ubiquitous Computing (ubicomp). The dominant technical approach in HAR includes sliding window segmentation of time-series data captured with body-worn sensors, manually designed feature extraction procedures, and a wide variety of (supervised) classification methods \cite{bulling2014tutorial}. In many cases, these relatively simple means suffice to obtain impressive recognition accuracies. However, more elaborate behaviours which are, for example, of interest in medical applications, pose a significant challenge to this manually tuned approach \cite{hammerla2015pd}. Some work has furthermore suggested that the dominant technical approach in HAR benefits from biased evaluation settings \cite{hammerla2015let}, which may explain some of the apparent inertia in the field of the adoption of deep learning techniques.

Deep learning has the potential to have significant impact on HAR in ubicomp. It can substitute for manually designed feature extraction procedures which lack the robust physiological basis that benefits other fields such as speech recognition. However, for the practitioner it is difficult to select the most suitable deep learning method for their application. Work that promotes deep learning almost always provides only the performance of the best system, and rarely includes details on how its seemingly optimal parameters were discovered. As only a single score is reported it remains unclear how this peak performance compares with the average during parameter exploration.

%Yet, possible benefits of deep learning in ubicomp has been identified by a variety of authors \cite{}. Deep learning has the potential to significantly improve the robustness of HAR systems in ubicomp by reducing the reliance on intuition and experience in feature selection \cite{}, by reducing the risk of overfitting to artificial study surroundings \cite{}, and through a more data-driven research methodology \cite{}.

In this paper we provide the first unbiased and systematic exploration of the peformance of state-of-the-art deep learning approaches on three different recognition problems typical for HAR in ubicomp. The training process for deep, convolutional, and recurrent models is described in detail, and we introduce a novel approach to regularisation for recurrent networks. In more than 4,000 experiments we investigate the suitability of each model for different tasks in HAR, explore the impact each model's hyper-parameters have on performance, and conclude guidelines for the practitioner who wants to apply deep learning to their application scenario. Over the course of these experiments we discover that recurrent networks outperform the state-of-the-art and that they allow novel types of real-time application of HAR through sample-by-sample prediction of physical activities.

\section{Deep Learning in Ubiquitous Computing}
\begin{figure*}[t]
	\centering
	\includegraphics[width=0.65\textwidth]{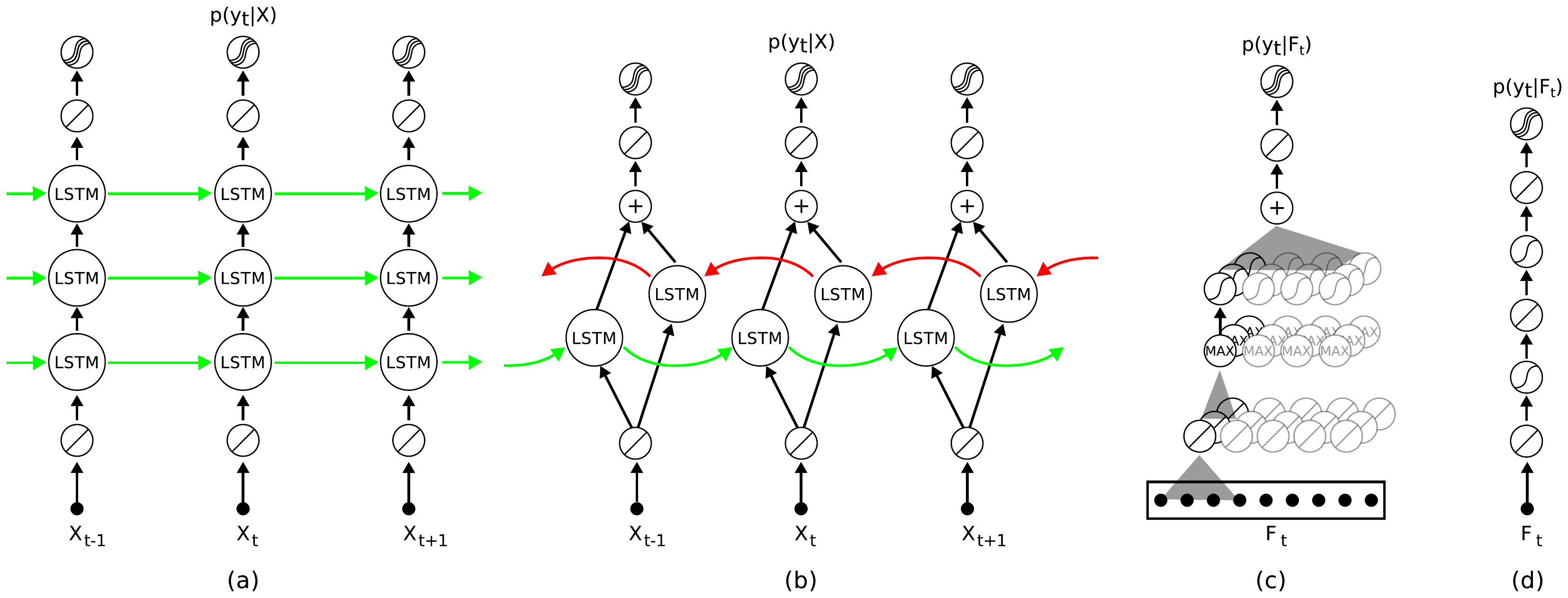}
	\caption{\label{fig:models}
	Models used in this work. From left to right: (a) LSTM network hidden layers containing LSTM cells and a final softmax layer at the top. (b) bi-directional LSTM network with two parallel tracks in both future direction (green) and to the past (red). (c) Convolutional networks that contain layers of convolutions and max-pooling, followed by fully-connected layers and a softmax group. (d) Fully connected feed-forward network with hidden (ReLU) layers.
	}
\end{figure*}

Movement data collected with body-worn sensors are multi-variate time-series data with relatively high spatial and temporal resolution (e.g. 20Hz - 200Hz). Analysis of this data in ubicomp is typically following a pipeline-based approach \cite{bulling2014tutorial}. The first step is to segment the time-series data into contiguous segments (or \textit{frames}), either driven by some signal characteristics such as signal energy \cite{plotz2012automatic}, or through a sliding-window segmentation approach. From each of the frames a set of features is extracted, which most commonly include statistical features or stem from the frequency domain.% \cite{figo2010preprocessing}.

The first deep learning approach explored to substitute for this manual feature selection corresponds to deep belief networks as auto-encoders, trained generatively with Restricted Boltzmann Machines (RBMs) \cite{plotz2011feature}. Results were mixed, as in most cases the deep model was outperformed by a combination of principal component analysis and a statistical feature extraction process. Subsequently a variety of projects have explored the use of pre-trained, fully-connected networks for automatic assessment in Parkinson's Disease \cite{hammerla2015pd}, as emission model in Hidden Markov Models (HMMs) \cite{zhang2015human,alsheikh2015deep}, and to model audio data for ubicomp applications \cite{lane2015deepear}.

The most popular approach so far in ubicomp relies on convolutional networks. Their performance for a variety of activity recognition tasks was explored by a number of authors \cite{zeng2014convolutional,ronao2015deep,yang2015deep,ronaoo2015evaluation}. Furthermore, convolutional networks have been applied to specific problem domains, such as the detection of stereotypical movements in Autism \cite{rad2015convolutional}, where they significantly improved upon the state-of-the-art.

Individual frames of movement data in ubicomp are usually treated as statistically independent, and applications of sequential modelling techniques like HMMs are rare \cite{bulling2014tutorial}. However, approaches that are able to exploit the temporal dependencies in time-series data appear as the natural choice for modelling human movement captured with sensor data. Deep recurrent networks, most notably those that rely on Long Short-Term Memory cells (LSTMs) \cite{hochreiter1997long}, have recently achieved impressive performance across a variety of scenarios (e.g.\ \cite{gregor2015draw}). Their application to HAR has been explored in two settings. First, \cite{neverova2015learning} investigated a variety of recurrent approaches to identify individuals based on movement data recorded from their mobile phone in a large-scale dataset. Secondly, \cite{ordonez2016deep} compared the performance of a recurrent approach to CNNs on two HAR datasets, representing the current state-of-the-art performance on Opportunity, one of the datasets utilised in this work. In both cases, the recurrent network was paired with a convolutional network that encoded the movement data, effectively employing the recurrent network to only model temporal dependencies on a more abstract level. Recurrent networks have so far not been applied to model movement data at the lowest possible level, which is the sequence of individual samples recorded by the sensor(s).

\section{Comparing deep learning for HAR}
While there has been some exploration of deep models for a variety of application scenarios in HAR there is still a lack of a systematic exploration of deep learning capabilities. Authors report to explore the parameter space in preliminary experiments, but usually omit the details. The overall process remains unclear and difficult to replicate. Instead, single instantiations of e.g.\ CNNs are presented that show good performance in an application scenario. Solely reporting peak performance figures does, however, not reflect the overall suitability of a method for HAR in ubicomp, as it remains unclear how much effort went into tuning the proposed approach, and how much effort went into tuning other approaches it was compared to. How likely is a practitioner to find a parameter configuration that works similarly well for their application? How representative is the reported performance across the models compared during parameter exploration? Which parameters have the largest impact on performance? These questions are important for the practitioner, but so far remain unanswered in related work.

In this paper we provide the first unbiased comparison of the most popular deep learning approaches on three representative datasets for HAR in ubicomp. They include typical application scenarios like manipulative gestures, repetitive physical activities, and a medical application of HAR in Parkinson's disease. We compare three types of models that are described below. To explore the suitability of each method we chose reasonable ranges for each of their hyperparameters and randomly sample model configurations. We report on their performance across thousands of experiments and analyse the impact of hyperparameters for each approach\footnote{Source-code will be made publicly available}.

\subsection{Deep feed-forward networks (DNN)}
We implemented deep feed-forward networks, which correspond to a neural network with up to five hidden layers followed by a softmax-group. The DNN represents a sequence of non-linear transformations to the input data of the network. We follow convention and refer to a network with N hidden layers as N-layer network. Each hidden layer contains the same number of units, and corresponds to a linear transformation and a recitified-linear (ReLU) activation function. We use two different regularisation techniques: i) Dropout: during training each unit in each hidden layer is set to zero with a probability $p_{drop}$, and during inference the output of each unit is scaled by $1/p_{drop}$ \cite{srivastava2014dropout} (dropout-rate is fixed to $0.5$ for all experiments); ii) Max-in norm: After each mini-batch the incoming weights of each unit in the network are scaled to have a maximum euclidean length of $d_{in}$. To limit the number of hyperparameters of the approach we chose not to perform any generative pre-training and to solely rely on a supervised learning approach. The input data fed into the network corresponds to frames of movement data. Each frame consists of a varying number of $s$ samples from $\mathbb{R}^d$, which are simply concatenated into a single vector $F_t \in \mathbb{R}^{s*d}$. The model is illustrated in figure \ref{fig:models}(d).

The DNN is trained in a mini-batch approach, where each mini-batch contains 64 frames and is stratified with respect to the class distribution in the training-set. We minimise the negative log likelihood using stochastic gradient descent.

\subsection{Convolutional networks (CNN)}
CNNs aim to introduce a degree of locality in the patterns matched in the input data and to enable translational invariance with respect to the precise location (i.e. time of occurrence) of each pattern within a frame of movement data. We explore the performance of convolutional networks and follow suggestions by \cite{srivastava2014dropout} in architecture and regularisation techniques. The overall CNN architecture is illustrated in figure \ref{fig:models}(c). Each CNN contains at least one temporal convolution layer, one pooling layer and at least one fully connected layer prior to a top-level softmax-group. The temporal convolution layer corresponds to a convolution of the input sequence with $n_f$ different \textit{kernels} (feature maps) of width $k_w$. Subsequent max-pooling is looking for the maximum within a region of width $m_w$ and corresponds to a subsampling, introducing translational invariance to the system. Width of the max-pooling was fixed to $2$ throughout all experiments. The output of each max-pooling layer is transformed using a ReLU activation function. The subsequent fully connected part effectively corresponds to a DNN and follows the same architecture outlined above.

For regularisation we apply dropout after each max-pooling or fully-connected layer, where the dropout-probability $p^i_{drop}$ in layer $i$ is fixed for all experiments ($p^1_{drop}=0.1$, $p^2_{drop}=0.25$, $p^{i>2}_{drop}=0.5$). Similar to the DNN we also perform max-in norm regularisation as suggested in \cite{srivastava2014dropout}. The input data fed into the CNN corresponds to frames of movement data as in the DNN. However, instead of concatenating the different input dimensions the matrix-structure is retained ($F_t \in \mathbb{R}^s \times \mathbb{R}^d$). The CNN is trained using stratified mini-batches (64 frames) and stochastic gradient descent to minimise negative log likelihood.

\subsection{Recurrent networks}
In order to exploit the temporal dependencies within the movement data we implemented recurrent neural networks based on LSTM cells in their \textit{vanilla} variant that does not contain \textit{peephole} connections \cite{greff2015lstm}. This architecture is recurrent as some of the connections within the network form a directed cycle, where the current timestep $t$ considers the states of the network in the previous timestep $t-1$. LSTM cells are designed to counter the effect of diminishing gradients if error derivatives are backpropagated through many layers ``through time" in recurrent networks \cite{hochreiter2001gradient}. Each LSTM cell (unit) keeps track of an internal state (the \textit{constant error carousel}) that represents it's ``memory". Over time the cells learn to output, overwrite, or null their internal memory based on their current input and the history of past internal states, leading to a system capable of retaining information across hundreds of time-steps \cite{hochreiter1997long}.

We implement two flavours of LSTM recurrent networks: i) deep forward LSTMs contain multiple layers of recurrent units and are connected ``forward" in time (see figure \ref{fig:models}(a)); and ii) bi-directional LSTMs which contain two parallel recurrent layers that stretch both into the ``future" and into the ``past" of the current time-step, followed by a layer that concatenates their internal states for timestep $t$ (see figure \ref{fig:models}(b)).

Practically these two flavours differ significantly in their application requirements. A forward LSTM contextualises the current time-step based on those it has seen previously, and is inherently suitable for real-time applications where, at inference time, the ``future" is not yet known. Bi-directional LSTMs on the other hand use both the future and past context to interpret the input at timestep $t$, which makes them suitable for offline analysis scenarios.
%Considering both future and past contexts to improve prediction is a concept popular in speech- or handwriting recognition and is exploited to train HMM-based language models \cite{}.

In this work we apply recurrent networks in three different settings, each of which is trained to minimise the negative log likelihood using \textit{adagrad} \cite{duchi2011adaptive} and subject to max-in norm regularisation.

In the first case the input data fed into the network at any given time $t$ corresponds to the current frame of movement data, which stretches a certain length of time and whose dimensions have been concatenated (as in the DNN above). We denote this model as \textit{LSTM-F}. The second application case of forward LSTMs represents a real-time application, where each sample of movement data is presented to the network in the sequence they were recorded, denoted \textit{LSTM-S}. The final scenario sees the application of bi-directional LSTMs to the same sample-by-sample prediction problem, denoted \textit{b-LSTM-S}.

\subsection{Training RNNs for HAR}
Common applications for RNNs include speech recognition and natural language processing. In these settings the context for an input (e.g. a word) is limited to it's surrounding entities (e.g.\ a sentence, paragraph). Training of RNNs usually treats these contextualised entities as a whole, for example by training an RNN on complete sentences.

In HAR the context of an individual sample of movement data is not well defined, at least beyond immediate correlations between neighbouring samples, and likely depends on the type of movement and its wider behavioural context. This is a problem well known in the field, and affects the choice of window length for sliding window segmentation \cite{bulling2014tutorial}.
%It makes it more difficult to decide on possible boundaries were a sequence of movement data could be "cut" and fed into an RNN as a self-contained unit.

In order to construct $b$ mini-batches that are used to train the RNN we initialise a number of positions $(p_i)_b$ between the start and end of the training-set. To construct a mini-batch we extract the $L$ samples that follow each position in $(p_i)_b$, and increase $(p_i)_b$ by $L$ steps, possibly wrapping around the end of the sequence. We found that it is important to initialise the positions randomly to avoid oscillations in the gradients. While this approach retains the ordering of the samples presented to the RNN it does not allow for stratification of each mini-batch w.r.t.\ class-distribution.

Training on long sequences has a further issue that is addressed in this work. If we use the approach outlined above to train a sufficiently large RNN it may ``memorise" the entire input-output sequence implicitly, leading to poor generalisation performance. In order to avoid this memorisation we need to introduce ``breaks" where the internal states of the RNN are reset to zero: after each mini-batch we decide to retain the internal state of the RNN with a carry-over probability $p_\text{carry}$, and reset it to zero otherwise. This is a novel form of regularisation of RNNs, which should be useful for similar applications of RNNs.

\section{Experiments}
The different hyper-parameters explored in this work are listed in table \ref{tab:hyper}. The last column indicates the number of parameter configurations sampled for each dataset, selected to represent an equal amount of computation time. We conduct experiments on three benchmark datasets representative of the problems tyical for HAR (described below). Experiments were run on a machine with three GPUs (NVidia GTX 980 Ti), where two model configurations are run on each GPU except for the largest networks.

After each epoch of training we evaluate the performance of the model on the validation set. Each model is trained for at least 30 epochs and for a maximum of 300 epochs. After 30 epochs, training stops if there is no increase in validation performance for 10 subsequent epochs. We select the epoch that showed the best validation-set performance and apply the corresponding model to the test-set.

\begin{table*}[]
\centering
\scriptsize
\begin{tabular}{ lr  ccc  ccc  ccccccccc  c }
 &
 & \rotatebox[origin=l]{90}{LR}
 & \rotatebox[origin=l]{90}{LR decay}
 & \rotatebox[origin=l]{90}{$L$}
  & \rotatebox[origin=l]{90}{momentum}
 & \rotatebox[origin=l]{90}{max-in norm}
 & \rotatebox[origin=l]{90}{$p_{\text{carry}}$}
 & \rotatebox[origin=l]{90}{\#layers}
 & \rotatebox[origin=l]{90}{\#units}
  & \rotatebox[origin=l]{90}{\#conv.-layers}
 & \rotatebox[origin=l]{90}{kW 1}
 & \rotatebox[origin=l]{90}{kW 2}
 & \rotatebox[origin=l]{90}{kW 3}
 & \rotatebox[origin=l]{90}{nF 1}
 & \rotatebox[origin=l]{90}{nF 2}
 & \rotatebox[origin=l]{90}{nF 3}
 & \rotatebox[origin=l]{90}{\#experiments} \\
\cline{3-17}
& Category & \multicolumn{3}{|c|}{Learning} & \multicolumn{3}{c|}{Regularisation} & \multicolumn{9}{c|}{Architecture} &  \\
\cline{3-17}
& log-uniform? & y & y & - & - &- &- &- &- &- &- &- &- &- &- &- &  \\
\hline
\multirow{2}{*}{DNN} & max & $10^{-1}$ & $10^{-3}$ & - & 0.99  & 4.0 & - & 5 & 2048 & - & - & - & - & - & - & - & \multirow{ 2}{*}{1000} \\
				& min & $10^{-4}$ & $10^{-5}$ & - & 0.0 & 0.5 & - & 1 & 64 & - & - & - & - & - & - & - \\
\hline
\multirow{2}{*}{CNN} & max & $10^{-1}$  & $10^{-3}$ & - & 0.99 & 4.0 & - & 3 & 2048 & 3 & 9 & 5 & 3 & 128 & 128 & 128 & \multirow{ 2}{*}{256}\\
				& min & $10^{-4}$  & $10^{-5}$ & - & 0.0 & 0.5 & - & 1 & 64 & 1 & 3 & 3 & 3 & 16 &  16 & 16 \\
\hline
\multirow{2}{*}{LSTM-F} & max & $10^{-1}$ & - & 64 & - & 4.0 & 1.0 & 3 & 384 & - & - & - & - & - & - & - & \multirow{ 2}{*}{128} \\
				  & min & $10^{-3}$  & - & 8 & - & 0.5 & 0.0 & 1 & 64 & - & - & - & - & - &  - & -  \\
\hline
\multirow{2}{*}{LSTM-S} & max & $10^{-1}$ & - & 196 & - & 4.0 & 1.0 & 3 & 384 & - & - & - & - & - & - & - & \multirow{ 2}{*}{128} \\
				  & min & $10^{-3}$ & - & 32 & - & 0.5 & 0.0 & 1 & 64 & - & - & - & - & - &  - & -  \\
\hline
\multirow{2}{*}{b-LSTM-S} & max & $10^{-1}$ & - & 196 & - & 4.0 & 1.0 & 1 & 384 & - & - & - & - & - & - & - & \multirow{ 2}{*}{128} \\
				     & min & $10^{-3}$ & - & 32 & - & 0.5 & 0.0 & 1 & 64 & - & - & - & - & - &  - & -  \\
\end{tabular}
\caption{Hyper-parameters of the models and the ranges of values explored in experiments.}
\label{tab:hyper}
\end{table*}

\subsection{Datasets}
We select three datasets typical for HAR in ubicomp for the exploration in this work. Each dataset corresponds to an application of HAR. The first dataset, Opportunity, contains manipulative gestures like opening and closing doors, which are short in duration and non-repetitive. The second, PAMAP2, contains prolonged and repetitive physical activities typical for systems aiming to characterise energy expenditure. The last, Daphnet Gait, corresponds to a medical application where participants exhibit a typical motor complication in Parkinson's disease that is known to have a large inter-subject variability. Below we detail each dataset:

The \textbf{Opportunity dataset (Opp)} \cite{chavarriaga2013opportunity} consists of annotated recordings from on-body sensors from 4 participants instructed to carry out common kitchen activities.
%Activities included preparing coffee, preparing a sandwich, moving objects, and putting objects back in their place or in the dishwasher after use.
Data is recorded at a frequency of 30Hz from 12 locations on the body, and annotated with 18 mid-level gesture annotations (e.g.\ Open Door / Close Door).
 %(Open Fridge, Move Cup, Open Door, Close Door, etc).
For each subject, data from 5 different runs is recorded.
We used the subset of sensors that did not show any packet-loss, which included accelerometer recordings from the upper limbs, the back, and complete IMU data from both feet.
%From the sensor data in the Opportunity Challenge [CITE], we used the subset of sensors which did not show any packet-loss. This included accelerometer data from upper and lower left and right arms, the back, in addition to 3D accelerometer, 3D rate of turn, 3D magnetic field and 3D orientation data from the left and right shoes.
The resulting dataset had 79 dimensions. We use run 2 from subject 1 as our validation set, and replicate the most popular recognition challenge by using runs 4 and 5 from subject 2 and 3 in our test set. The remaining data is used for training. For frame-by-frame analysis, we created sliding windows of duration 1 second and 50\% overlap. The resulting training-set contains approx.\ $650k$ samples ($43k$ frames).

The \textbf{PAMAP2 dataset} \cite{reiss2012introducing} consists of recordings from 9 participants instructed to carry out 12 lifestyle activities, including household activities and a variety of exercise activities (Nordic walking, playing soccer, etc). Accelerometer, gyroscope, magnetometer, temperature and heart rate data are recorded from inertial measurement units located on the hand, chest and ankle over 10 hours (in total). The resulting dataset has 52 dimensions.
We used runs 1 and 2 for subject 5 in our validation set and runs 1 and 2 for subject 6 in our test set. The remaining data is used for training.
In our analysis, we downsampled the accelerometer data to 33.3Hz in order to have a temporal resolution comparable to the Opportunity dataset. For frame-by-frame analysis, we replicate previous work with non-overlapping sliding windows of 5.12 seconds duration with one second stepping between adjacent windows (78\% overlap) \cite{reiss2012introducing}. The training-set contains approx.\ $473k$ samples ($14k$ frames).

The \textbf{Daphnet Gait dataset (DG)} \cite{bachlin2009potentials} consists of recordings from 10 participants affected with Parkinson's Disease (PD), instructed to carry out activities which are likely to induce freezing of gait. Freezing is a common motor complication in PD, where affected individuals struggle to initiate movements such as walking. The objective is to detect these freezing incidents, with the goal to inform a future situated prompting system. This represents a two-class recognition problem. Accelerometer data was recorded from above the ankle, above the knee and on the trunk. The resulting dataset has 9 dimensions.
We used run 1 from subject 9 in our validation set, runs 1 and 2 from subject 2 in our test set, and used the rest for training.
In our analysis, we downsampled the accelerometer data to 32Hz. For frame-by-frame analysis, we created sliding windows of 1 second duration and 50\% overlap. The training-set contains approx.\ $470k$ samples ($30k$ frames).

%\begin{table}[b]
%\centering
%\caption{Data-set properties}
%\label{tab:datasets}
%\begin{tabular}{lllll}
 %            & \#samples & \textit{f} & \#frames & \#classes \\
%Opportunity  &  $118,750$ & $30$Hz & $43,397$ & $18$          \\
%PAMAP2       &           & $33$Hz &          &           \\
%Daphnet Gait &           & $32$Hz &          &
%\end{tabular}
%\end{table}

\subsection{Influence of hyper-parameters}
In order to estimate the impact of each hyperparameter on the performance observed across all experiments we apply the fANOVA analysis framework. fANOVA \cite{hutter2014efficient} determines the extent to which each hyperparameter contributes to a network's performance. It builds a predictive model (random forest) of the model performance as a function of the model's hyperparameters.
%Using performance information from experiments ran, a random forest is used to model the predictive performance in terms of the hyperparameters from each run.
This non-linear model is then decomposed into marginal and joint interaction functions of the hyperparameters, from which the percentage contribution to overall variability of network performance is obtained. fANOVA has been used previously to explore the hyperparameters in recurrent networks by \cite{greff2015lstm}.
%This indicates which hyperparameters are the most crucial to tune in order to build neural networks with better performance.

For the practitioner it is important to know which aspect of the model is the most crucial for performance. We grouped the hyperparameters of each model into one of three categories (see table \ref{tab:hyper}): i) \textit{learning}, parameters that control the learning process; ii) \textit{regularisation}, parameters that limit the modelling capabilities of the model to avoid overfitting; and iii) \textit{architecture}, parameters that affect the structure of the model. Based on the variability observed for each hyperparameter we estimate the variability that can be attributed to each parameter category, and to higher order interactions between categories.

\subsection{Performance metrics}
As the datasets utilised in this work are highly biased we require performance metrics that are independent of the class distribution. We opted to estimate the mean f1-score:
%, which corresponds to the harmonic mean of precision and recall:
\begin{equation}
F_m = \frac{2}{|c|} \sum_c \frac{\text{prec}_c \times \text{recall}_c}{\text{prec}_c + \text{recall}_c}
\end{equation}
Related work has previously used the weighted f1-score as primary performance metric (for Opportunity). In order to compare our results to the state-of-the-art we estimate the weighted f1-score:
% for the best performing models discovered in our experiments:
\begin{equation}
F_w = 2 \sum_c  \frac{N_c}{N_{\text{total}}} \frac{\text{prec}_c \times \text{recall}_c}{\text{prec}_c + \text{recall}_c},
\end{equation}
where $N_c$ is the number of samples in class $c$, and $N_{\text{total}}$ is the total number of samples.

\begin{table}[]
\centering
\scriptsize
\begin{tabular}{ l  cccc }
& PAMAP2 & DG  & \multicolumn{2}{c}{OPP} \\
Performance & $F_m$ & $F_1$ & $F_m$ & $F_w$  \\
\hline
DNN & $0.904$ & $0.633$ & $0.575$ & $0.888$  \\
CNN & $\mathbf{0.937}$ & $0.684$ & $0.591$ & $0.894$ \\
LSTM-F & $0.929$ & $0.673$& $0.672$ & $0.908$  \\
LSTM-S & $0.882$ & $\mathbf{0.760}$& $0.698$ & $0.912$  \\
b-LSTM-S & $0.868$ & $0.741$ & $\mathbf{0.745}$ & $\mathbf{0.927}$  \\
\hline
CNN & \multicolumn{2}{l}{\cite{yang2015deep}} & $-$ & $0.851$ \\
CNN & \multicolumn{2}{l}{\cite{ordonez2016deep}} & $0.535$ & $0.883$ \\
DeepConvLSTM & \multicolumn{2}{l}{\cite{ordonez2016deep}} & $0.704$ & $0.917$ \\
\hline
Delta from median & $\Delta F_m$ & $\Delta F_1$ & $\Delta F_m$ & mean \\
DNN & $0.129$ & $0.149$ & $0.357$ & $0.221$  \\
CNN & $\mathbf{0.071}$ & $\mathbf{0.122}$ & $0.120$ & $\mathbf{0.104}$ \\
LSTM-F & $0.10$ & $0.281$& $0.085$ & $0.156$  \\
LSTM-S & $0.128$ & $0.297$& $\mathbf{0.079}$ & $0.168$  \\
b-LSTM-S & $0.087$ & $0.221$ & $0.205$ & $0.172$ \\
\end{tabular}
\caption{Best results obtained for each model and dataset, along with some baselines for comparison. Delta from median (lower part of table) refers to the absolute difference between peak and median performance across all experiments.}
\label{tab:results}
\end{table}

\section{Results}
\begin{figure*}[t]
	\centering
	\includegraphics[width=\textwidth]{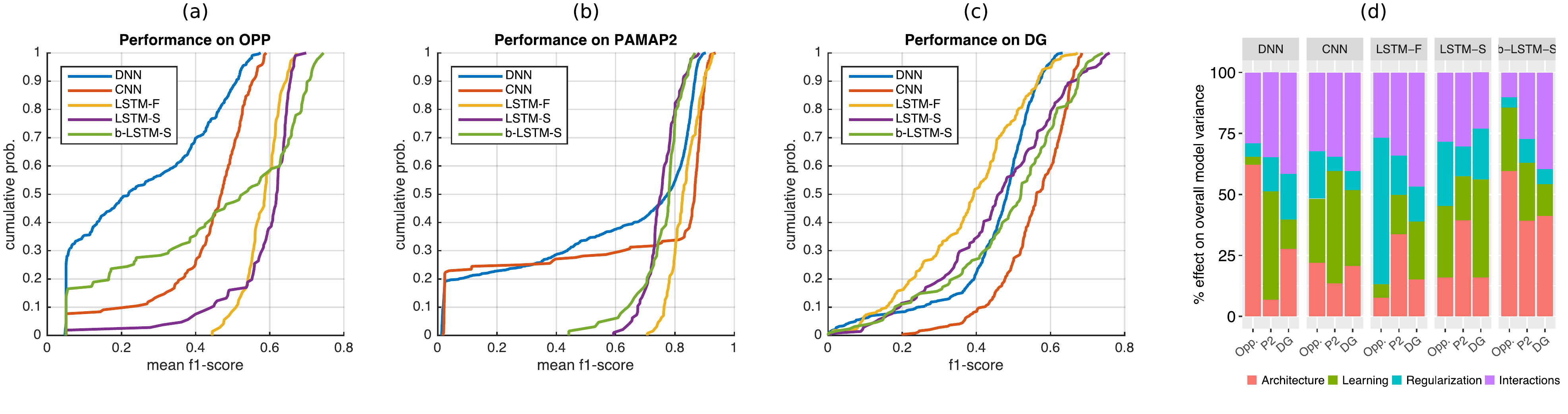}
	\caption{\label{fig:results}
	(a)-(c): Cumulative distribution of recognition performance for each dataset. (d): results from fANOVA analysis, illustrating impact of hyperparameter-categories on recognition performance (see table \ref{tab:hyper}).
	}
\end{figure*}

%Results across all experiments are illustrated in figure \ref{fig:results}. Graphs (a) to (c) show the cumulative distribution of the mean f1-score (OPP, PAMAP2), or of the f1-score for the objective class (DG), across all the models utilised in this work. Graph (d) illustrates the importance of each group of hyperparameters estimated using fANOVA. Table \ref{tab:results} lists the best performing systems and baseline scores of other deep approaches on Opportunity. Furthermore it lists the difference between the best performing model and the median model for each setting.

Results are illustrated in figure \ref{fig:results}. Graphs (a-c) show the cumulative distribution of the main performance metric on each dataset. Graph (d) illustrates the effect of each category of hyper-parameter estimated using fANOVA.

Overall we observe a large spread of peak performances between models on OPP and DG, with more than 15\% mean f1-score between the best performing approach (b-LSTM-S) and the worst (DNN) on OPP (12\% on DG) (see table \ref{tab:results}). On PAMAP2 this difference is smaller, but still considerable at 7\%. The best performing approach on OPP (b-LSTM-S) outperforms the current state-of-the-art by a considerable margin of 4\% mean f1-score (1\% weighted f1-score). The best CNN discovered in this work further outperforms previous results reported in the literature for this type of model by more than 5\% mean f1-score and weighted f1-score (see table \ref{tab:results}). The good performance of recurrent approaches, which model movement at the sample level, holds the potential for novel (real-time) applications in HAR, as they alleviate the need for segmentation of the time-series data.

The distributions of performance scores differ between the models investigated in this work.
%The models explored in this work differ in how their performance is distributed.
CNNs show the most characteristic behaviour: a fraction of model configurations do not work at all (e.g.\ 20\% on PAMAP2), while the remaining configurations show little variance in their performance. On PAMAP2, for example, the difference between the peak and median performance is only 7\% mean f1-score (see table \ref{tab:results}). The DNNs show the largest spread between peak and median performance of all approaches of up to 35.7\% on OPP. Both forward RNNs (LSTM-F, LSTM-S) show similar behaviour across the different datasets. Practically all of their configurations explored on PAMAP2 and OPP have non-trivial recognition performance.
%, showing slightly higher mean difference between peak and median scores.

The effect of each category of hyperparameter on the recognition performance is illustrated in figure \ref{fig:results}(d). Interestingly, we observe the most consistent effect of the parameters in the CNN. In contrast to our expectation it is the parameters surrounding the learning process (see table \ref{tab:hyper}) that have the largest main effect on performance. We expected that for this model the rich choice of architectural variants should have a larger effect. For DNNs we do not observe a systematic effect of any category of hyperparameter. On PAMAP2, the correct learning parameters appear to the be the most crucial. On OPP it is the architecture of the model. Interestingly we observed that relatively shallow networks outperform deeper variants. There is a drop in performance for networks with more than 3 hidden layers. This may be related to our choice to solely rely on supervised training, where a generative pre-training may improve the performance of deeper networks.

The performance of the frame-based RNN (LSTM-F) on OPP depends critically on the  carry-over probability introduced in this work. Both always retaining the internal state and always forgetting the internal state lead to the low performance. We found that $p_\text{carry}$ of 0.5 works well for most settings. Our findings merit further investigation, for example into a carry-over \textit{schedule}, which may further improve LSTM performance.

Results for sample-based forward LSTMs (LSTM-S) mostly confirm earlier findings for this type of model that found learning-rate to be the most crucial parameter \cite{greff2015lstm}. However, for bi-directional LSTMs (b-LSTM-S) we observe that the number of units in each layer has a suprisingly large effect on performance, which should motivate practitioners to first focus on tuning this parameter.

\section{Discussion}
In this work we explored the performance of state-of-the-art deep learning approaches for Human Activity Recognition using wearable sensors. We described how to train recurrent approaches in this setting and introduced a novel regularisation approach. In thousands of experiments we evaluated the performance of the models with randomly sampled hyper-parameters. We found that bi-directional LSTMs outperform the current state-of-the-art on Opportunity, a large benchmark dataset, by a considerable margin.

However, interesting from a practitioner's point of view is not the peak performance for each model, but the process of parameter exploration and insights into their suitability for different tasks in HAR. Recurrent networks outperform convolutional networks significantly on activities that are short in duration but have a natural ordering, where a recurrent approach benefits from the ability to contextualise observations across long periods of time. For bi-directional RNNs we found that the number of units per layer has the largest effect on performance across all datasets. For prolonged and repetitive activities like walking or running we recommend to use CNNs. Their average performance in this setting makes it more likely that the practitioner discovers a suitable configuration, even though we found some RNNs that work similarly well or even outperform CNNs in this setting. We further recommend to start exploring learning-rates, before optimising the architecture of the network, as the learning-parameters had the largest effect on performance in our experiments.

%Due to the higher average performance in this settings it is simply more likely that a practitioner discovers a well performing configuration, despite the fact that some recurrent approaches show similar performance or even outperform them.

We found that models differ in the spread of recognition performance for different parameter settings. Regular DNNs, a model that is probably the most approachable for a practitioner, requires a significant investment in parameter exploration and shows a substantial spread between the peak and median performance. Practitioners should therefore not discard the model even if a preliminary exploration leads to poor recognition performance. More sophisticated approaches like CNNs or RNNs show a much smaller spread of performance, and it is more likely to find a configuration that works well with only a few iterations.

\newpage{}
\bibliographystyle{named}
\bibliography{deep}

\end{document}